%% file: main.tex
\pdfoutput=1

\documentclass[11pt]{article}

\usepackage[final]{acl}

\usepackage{times}
\usepackage{latexsym}

\usepackage{graphicx}
\usepackage{multirow}
\usepackage{pgf-pie}   
\usepackage{xcolor} 
\usepackage{tablefootnote}
\usepackage{tabularx}
\usepackage{adjustbox}
\usepackage{float}
\usepackage{graphicx}
\usepackage{caption}
\usepackage{subcaption}
\usepackage{threeparttable}
\usepackage{lipsum}
\usepackage{flushend}
\usepackage{longtable}
\usepackage{amsmath}
\usepackage{makecell}

\usepackage{array,xcolor}

\usepackage{xcolor}
\usepackage{soul, soulutf8}

\usepackage[T5]{fontenc}
\usepackage{CJKutf8}

\usepackage{hyperref}

\title{EVJVQA Challenge: Multilingual Visual Question Answering}

\author{
Ngan Luu-Thuy Nguyen$^1$, Nghia Hieu Nguyen$^2$, Duong T. D. Vo$^3$, Khanh Quoc Tran$^4$, \\\bf Kiet Van Nguyen$^5$\\
Faculty of Information Science and Engineering, University of Information Technology, \\Ho Chi Minh City, Vietnam \\
Vietnam National University, Ho Chi Minh City, Vietnam \\
 {\tt \{19520178$^2$,19520483$^3$\}@gm.uit.edu.vn},\\ {\tt \{ngannlt$^1$,khanhtq$^4$,kietnv$^5$\}@uit.edu.vn} \\}

\begin{document}
\maketitle
\begin{abstract}

Visual Question Answering (VQA) is a challenging task of natural language processing (NLP) and computer vision (CV), attracting significant attention from researchers. English is a resource-rich language that has witnessed various developments in datasets and models for visual question answering. Visual question answering in other languages also would be developed for resources and models. In addition, there is no multilingual dataset targeting the visual content of a particular country with its own objects and cultural characteristics. To address the weakness, we provide the research community with a benchmark dataset named EVJVQA, including 33,000+ pairs of question-answer over three languages: Vietnamese, English, and Japanese, on approximately 5,000 images taken from Vietnam for evaluating multilingual VQA systems or models. EVJVQA is used as a benchmark dataset for the challenge of multilingual visual question answering at the 9th Workshop on Vietnamese Language and Speech Processing (VLSP 2022). This task attracted 62 participant teams from various universities and organizations. In this article, we present details of the organization of the challenge, an overview of the methods employed by shared-task participants, and the results. The highest performances are 0.4392 in F1-score and 0.4009 in BLUE on the private test set. The multilingual QA systems proposed by the top 2 teams use ViT for the pre-trained vision model and mT5 for the pre-trained language model, a powerful pre-trained language model based on the transformer architecture. EVJVQA is a challenging dataset that motivates NLP and CV researchers to further explore the multilingual models or systems for visual question answering systems. We released the challenge on the \href{https://codalab.lisn.upsaclay.fr/competitions/12274}{Codalab} evaluation system for further research.

\end{abstract}

\input{Sections/1-Introduction}
\input{Sections/2-Related-works}

\input{Sections/3-Task.tex}
\input{Sections/4-Datasets}
\input{Sections/5-Experiments}

\input{Sections/6-Result-Analysis}
\input{Sections/7-Conclusion}

\section*{Acknowledgements}

We would like to thank the efforts of annotators who have contributed to the construction of a high-quality resource for the natural language processing research community. The VLSP 2022 was supported by organizations: Aimsoft, Bee, INT2, and DAGORAS, and educational organizations: VNU-HCM University of Information Technology, VNU University of Science, VNU University of Engineering and Technology, Hanoi University of Science and Technology, Vietnam Lexicography Centre, University of Science and Technology of Hanoi, ThuyLoi University, and VAST Institute of Information Technology. This work is partially supported by Vingroup Innovation Foundation (VINIF) in project code VINIF.2020.DA14.
\bibliography{anthology}
\bibliographystyle{acl_natbib}


\appendix

\input{Sections/Appendix}

\end{document}

%% file: Sections/1-Introduction.tex
\section{Introduction}

Visual or Image-Based Question Answering is a challenging task that requires knowledge of two hot AI fields: natural language processing and computer vision. Specifically, querying the information of images through human-language questions is a friendly and natural approach to searching for information, meeting the needs of people extracting information in many domains such as life, education, work, etc. However, studies have mainly focused on resource-rich languages such as English. In this challenge, we aim to extend visual question answering to more languages, including rich and low-resource languages, including resources and models.

English has witnessed numerous benchmarks for evaluating and developing visual question answering models or systems. Recently, researchers designed  datasets with goal-oriented evaluations. Firstly, VQA datasets \cite{antol2015vqa,zhu2016visual7w,goyal2017making,changpinyo2022all} were created on general images. Soon after, the more complex dataset based on reasoning is discovered by \cite{hudson2019gqa}. In addition, VQA \cite{gurari2018vizwiz} is also geared towards support applications for seemingly-impaired and blind people. \cite{singh2019towards} showed that VQA has the ability to read text on photos. More challenging, VQA  requires external, commonsense, or world knowledge to predict more correct answers \cite{marino2019ok,schwenk2022okvqa}. Besides, Changpinyo et al. \cite{changpinyo2022towards} proposed a multilingual dataset for visual question answering in 13 languages. However, this dataset is done automatically based on the auto-translation and verification method. In this paper, we presented a human-generation multilingual dataset, including three languages: English (resource-rich language), Japanese, and Vietnamese(resource-poor language).

In this paper, we have three main contributions described as follows.
\begin{enumerate}
    \item Firstly, we constructed UIT-EVJVQA, a multilingual dataset for evaluating the visual question answering systems or models, which comprises 33.790 question-answer pairs in three languages: English, Vietnamese, and Japanese.
    
    \item We organize the VLSP2022-EVJVQA Challenge for evaluating multilingual VQA models (Vietnamese, English, and Japanese) at the VLSP 2022. Our baseline system obtains 0.3346 in F1-score and 0.2275 in BLEU on the public and private test sets, respectively, and there are no models of participating teams that pass 0.44 (in F1-score) on the private test set, which indicates our dataset is challenging and  requires the development of multilingual VQA models.
    
    \item When combined with other VQA datasets for analysis, UIT-EVJVQA could potentially be a useful resource for multilingual research.
\end{enumerate}

The following is how the rest of the article is organized. In Section 2, we provide a brief overview
of the background and relevant studies. We introduce the VLSP 2021-EVJVQA Challenge in Section 3. Our new dataset (UIT-EVJVQA) is presented in detail in Section 4. Section 5 presents the systems and results proposed by participating teams. In Section 6, we provide further analysis of the challenge results. Finally, Section 7 summarizes the findings of the VLSP 2022-EVJVQA Challenge and suggests future research directions.

%% file: Sections/2-Related-works.tex
\section{Background and Related Work}

Visual Question Answering (VQA) is a challenging task that has significant  value not only in the research community but also in daily life. VQA task was first introduced by \cite{antol2015vqa}. The authors were successful in creating a novel dataset and fundamental English methodologies. Inspired by that success, various further studies have been created and implemented in a variety of languages \cite{gupta2017survey} including Chinese \cite{qi2022dureadervis}, Japanese \cite{shimizu2018visual}, and Vietnamese \cite{tran-etal-2021-vivqa}.

VQA has gained more attention from researchers in recent years and has shown significant growth. Studies are currently being introduced not only in monolingual but also in multilingual applications \cite{pfeiffer2021xgqa,khan2021towards,liu2022delving, nooralahzadeh2022improving}. This stage contributes significantly to the creation of multilingual VQA (mVQA) systems. Some typical research works in this approach can be mentioned such as \cite{gupta2020unified} with the study that the proposed model is capable of predicting responses from questions in Hindi, English, or Code-mixed (Hinglish: Hindi-English) languages; Changpinyo et al. \cite{changpinyo2022towards} with a translation-based mVQA dataset in 7 distinct languages; Gao et al. \cite{gao2015you} construct a Freestyle Multilingual Image Question Answering (FM-IQA) dataset contains over 150,000 images and 310,000 freestyle Chinese question-answer pairs and their English translations.

In this study, we create the first dataset for the task of mVQA on English-Vietnamese-Japanese (EVJVQA). The EVJVQA dataset is expected to open up new research areas and aid in evaluating multilingual VQA models.

%% file: Sections/3-Task.tex
\section{The VLSP 2022 - EVJVQA Challenge}

\subsection{Task Definition}

\begin{figure*}
    \centering
    \includegraphics[width=0.9\textwidth]{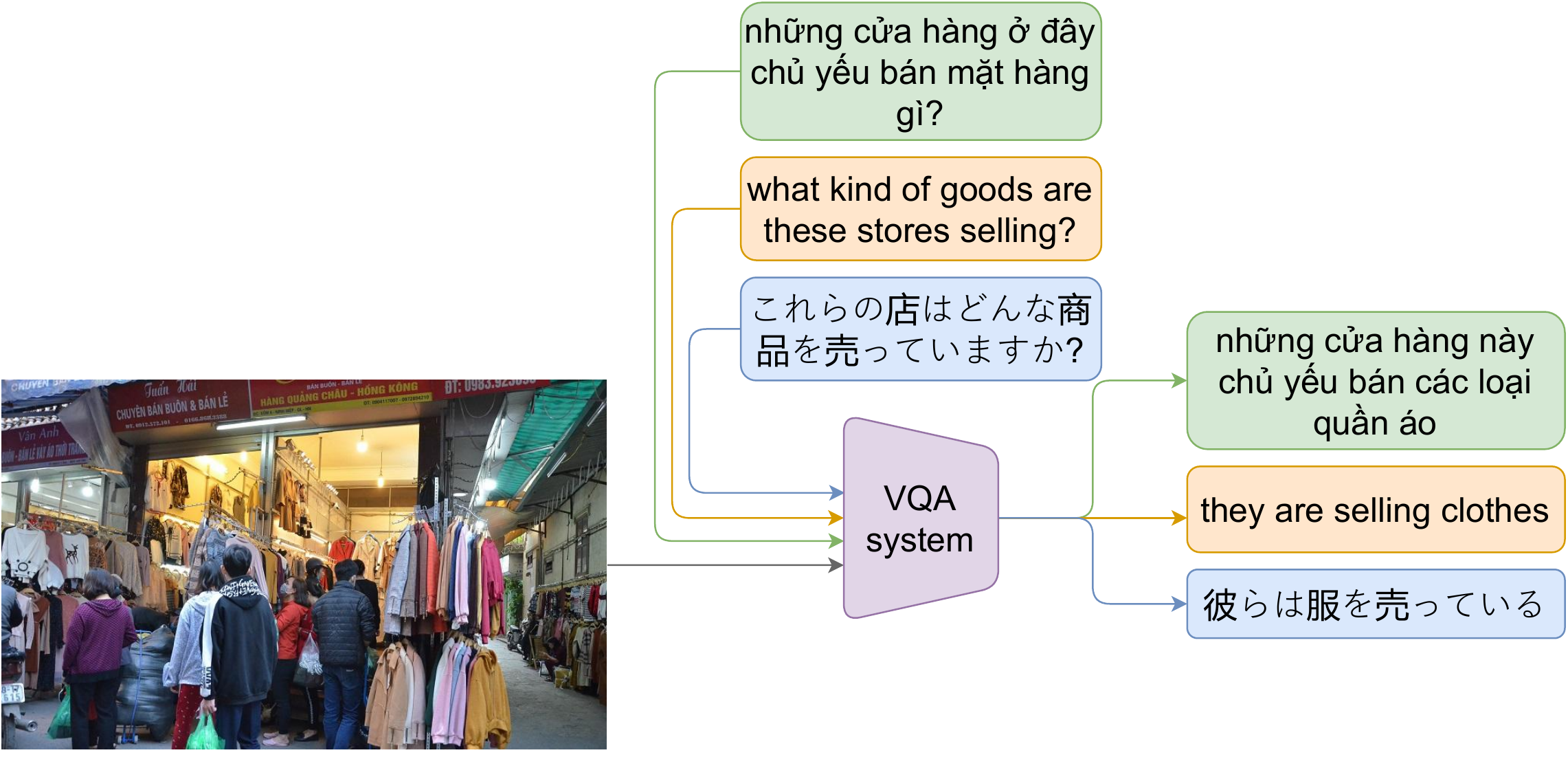}
    \caption{Overview of the multilingual visual question answering task.}
    \label{fig:task_def}
\end{figure*}

This task aims to enable the ability of computers to understand images and answer relevant questions in different languages from users. The task is defined as below (Figure \ref{fig:task_def}):

\begin{itemize}
    \item \textbf{Input}: Given an image and a question that can be answerable.
    \item \textbf{Output}: An answer where can be a span related to the image's content.
\end{itemize}

\textbf{Language Selection}. Three languages are selected in which the main language is Vietnamese, and the other two popular other languages (English and Japanese) in the pictures taken from Vietnam.


\subsection{Evaluation Metrics}

In this challenge, we use two evaluation metrics: F1 and BLUE. Based on \cite{rajpurkar2016squad}, the F1-score of each answer is calculated based on tokens of the gold answer (GA) and tokens of the predicted answer (PA). The overall F1 is averaged across all questions of each set. For Vietnamese and English languages, we calculate F1 based on tokens, whereas F1 is calculated based on characters for Japanese.

$$
    Precision (P) = \frac{GA \cap PA}{PA}
$$

\[
    Recall (R) = \frac{GA \cap PA}{GA}
\]

\[
    F1 = \frac{2PR}{P + R}
\]

Inspired by \cite{papineni2002bleu}, the \textbf{B}ilingual \textbf{E}valuation \textbf{U}nderstudy (BLEU), a popular evaluation metric in machine translation, computes the n-gram co-occurrence between human-generation answers and system-generation answers. The best performances were estimated by averaged BLEU-based performances (BLEU-1, BLEU-2, BLEU-3, and BLEU-4) on the public test and private test sets. Both evaluation metrics ignore punctuations.

\subsection{Schedule and Overview Summary}

Table \ref{importantdate} shows important dates of the VLSP 2022 - EVJVQA Challenge. It lasted for two months, during which the participating teams spent 39 days developing the multilingual visual question answering systems.

\begin{table}[ht]
\centering
\caption{Schedule of the VLSP 2021 - ViMRC Challenge.}
\begin{tabular}{l l}
\hline
\multicolumn{1}{c}{\textbf{Time}} & \multicolumn{1}{c}{\textbf{Phase}} \\ \hline
October 1st                         & Trial Data                          \\ 
October 5th                         & Public test                         \\ 
November 10th                        & Private test                        \\ 
November 12th                        & Competition end                     \\ \hline
November 20th                       & Submission deadline                 \\ 
November 23th                       & Notification of acceptance          \\ 
November 25th                       & Camera-ready due                    \\ \hline
\end{tabular}
\label{importantdate}
\end{table}

Besides, Table \ref{tbl_participate_sumary} presents an overview of the participating teams who joined the VLSP2022-EVJVQA.

\begin{table}[ht]
    \centering
    \caption{Participation summary of the VLSP 2022 - EVJVQA Challenge.}
    \begin{tabular}{l r}
        \hline
        \textbf{Metric} & \textbf{Value} \\
        \hline
        \#Registration Teams & 62 \\
        \#Joined Teams & 57 \\
        \#Signed Data Agreements & 36 \\
        \#Submitted Teams & 8 \\
        \#Paper Submissions & 5 \\
        \hline
    \end{tabular}
    
    \label{tbl_participate_sumary}
\end{table}

%% file: Sections/4-Datasets.tex
\section{Corpus Creation}

\begin{figure*}[ht]
    \centering
    \includegraphics[width=0.8\linewidth]{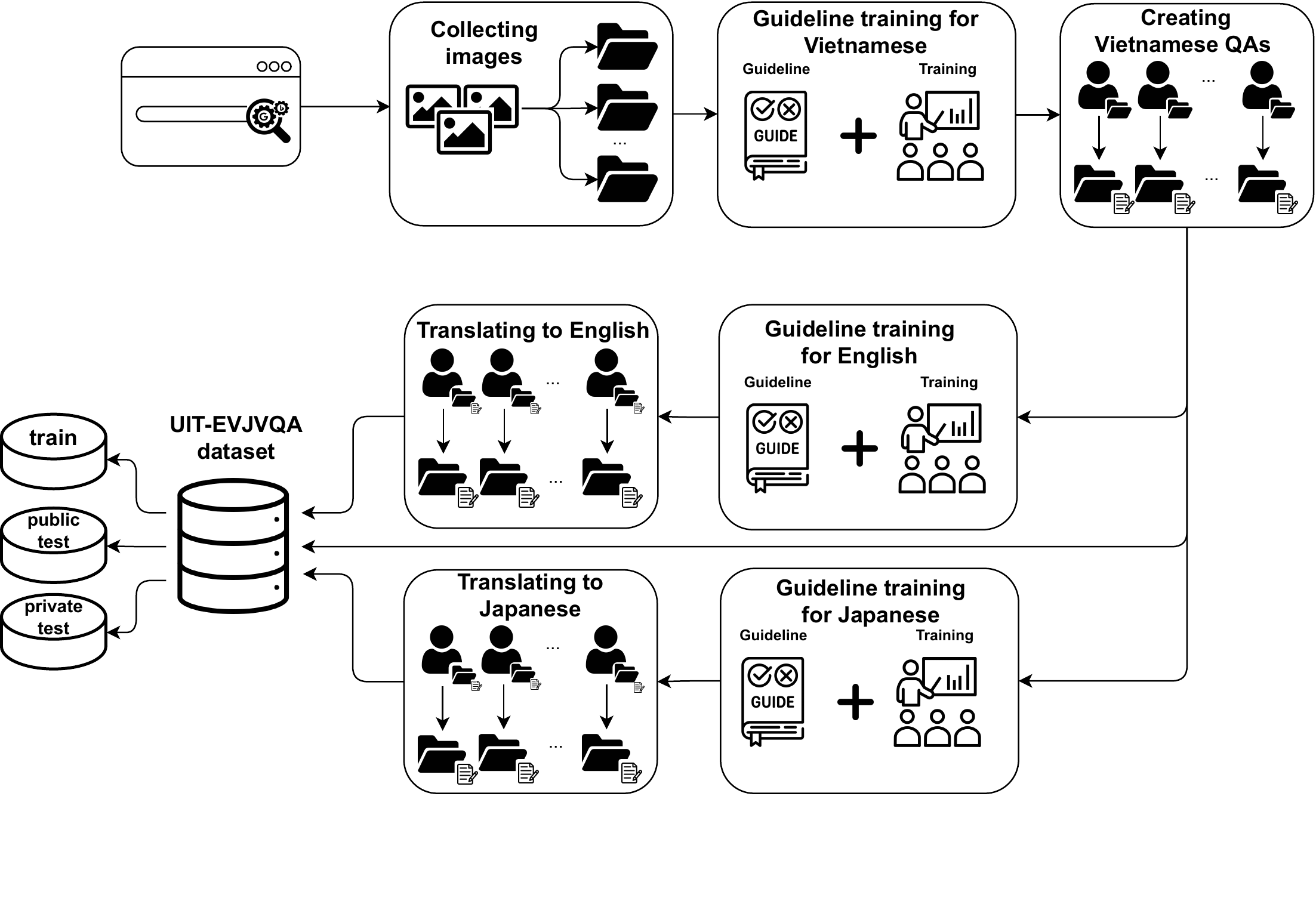}
    \caption{Overall pipeline process for creating the UIT-EVJVQA dataset.}
    \label{fig:pipeline-evjvqa}
\end{figure*}
A previous work \cite{tran2021vivqa} inherited assets from the well-known VQA benchmark in English and the COCO-QA \cite{ren2015exploring}, then they proposed a semi-automatic annotating system by using machine translation to translate question-answer pairs from English to Vietnamese. On the other hand, we argue that the context in images captured in Vietnam is more complicated than in images coming from VQA benchmarks in English because of its crowded scene and "out-of-common" objects, or in particular, objects that are not commonly used outside of Vietnam. Moreover, using such a machine translation system as \cite{tran2021vivqa} is hard to ensure the natural aspect of using language, which caused lots of confusion while evaluating VQA methods in Vietnamese. To overcome the above flaws and research and develop a VQA system, particularly for the Vietnamese, we constructed the novel dataset with images collected manually and relevant to daily life in Vietnam. In addition, to address and challenge the research community, we provided our dataset \textit{\textbf{multilingual}} question-answer pairs to encourage the research community to explore and propose an effective system that can answer questions written in various languages.

\subsection{Image collection}
To build a VQA dataset in the Vietnamese context, we search for images with a diverse and open-domain set of keywords. We first selectively prepare various keywords which are relevant to Vietnamese locations and daily life activities or result in images that specifically contain targeted objects in Vietnamese scenes. For instance, the keywords can be Vietnamese streets, markets, sidewalk eateries, cultural sites, human outdoor activities, means of transport, or house interiors. Some keywords are appended with Vietnamese location names like Hanoi or Saigon for more variation in geological and cultural context. We then use these keywords to scrap images from Google Images. The image-scraping process is facilitated by the Octoparse tool.

After collecting the images, we proceed with the filtering stage. The images originally came in various sizes. However, we must ensure the details in them are clearly visible. Therefore, we only keep images with widths and heights greater than 500 and 400, respectively. We also filter out GIF files and other file formats apart from JPEG and PNG. As a result, we obtain 4,909 images with their size varying in the range of 500 - 6000 pixels in terms of width and the range of 400 - 4032 pixels in terms of height.
\subsection{Questions and answers creation process}

The questions and answers (QAs) of the UIT-EVJVQA dataset are first created in Vietnamese throughout the set of images. Afterward, these QAs are translated into English and Japanese. Both stages are conducted by the source of crowd workers. QAs in all three languages are then merged according to the images and eventually constitute the final corpus. The overall pipeline of the aforementioned process is visualized in Figure \ref{fig:pipeline-evjvqa}.

\subsubsection{Base Vietnamese QAs creation} \label{guideline}
We first employ five crowd workers for the Vietnamese questions and answers creation stage.
For each image, the workers are asked to formulate 3-5 question-answer pairs based on the \textbf{details} and objects that appear in the visible scene. The workers are required to conform to the following guideline:
\begin{itemize}
    \item Encourage using phrases or full sentences to give the answers.
    \item Restrict the use of single words as answers.
    \item No selective question or yes/no question allowed.
    \item Numbers must be typed in alphabet characters rather than numeric characters, and they must not be greater than 10.
    \item For colors, only use provided colors such as black, white, red, orange, yellow, green, blue, pink, purple, brown, and grey. If these colors cannot exactly describe the true color of the object, ignore such color property in the sentence.
    \item In the case of mentioning direction, if following that direction words is an object, then such direction is defined based on that object, else using the perspective of the annotator to define the direction.
\end{itemize}

Eventually, this stage yields 11,689 pairs of question and answer in Vietnamese for our dataset.

\subsubsection{English and Japanese QAs human-translation}
All the Vietnamese QAs are henceforth passed through the human-translation stages. These stages demand the employment of qualified crowd workers for the translation of questions and answers. For English translation, the workers must have at least IELTS certification with an overall band score of 6.5. Meanwhile, translators for the Japanese translation task must achieve an N3 proficiency level or above. Overall, there 
were seven and nine translators working simultaneously for the English and Japanese translation process, respectively. The English and Japanese translation stages also have their corresponding guidelines.

\textbf{English translation guideline}: The English QAs are translated from the Vietnamese ones, with many entities and attributes in the sentences retained during the translation as possible. The translators are encouraged to use phrases or full sentences as the translated answers. Apostrophes in sentences are restricted. Thus, translators should use the uncontracted form or other valid grammar formulations.

\textbf{Japanese translation guideline}: The Japanese QAs are translated from the Vietnamese ones, with many entities and attributes in the sentences retained during the translation as possible. For transcribing Vietnamese proper nouns or other foreign words, the katakana syllabary is adopted. The polite form is utilized for writing translated questions and answers that contain verbs. In the case of complex Vietnamese questions and answers with multiple relevant information that may not be easily translated into continuous Japanese sentences, translators can use commas to split the sentences into smaller parts and then translate them subsequently. For example, the question "What products does the woman wearing a helmet go to the store to buy?" can be translated to Japanese as "\begin{CJK}{UTF8}{min}ヘルメットをかぶった女性が店に買いに行って、どうの商品を買いますか?\end{CJK}", which literally means "A woman wearing a helmet goes shopping at a store, what products does she buy?".  

After the human-translation process, we obtained 10,539 question-answer pairs in English and 11,562 question-answer pairs in Japanese.

\subsection{Pre-processing and splitting}
We normalize the Vietnamese and English QAs into lowercase. Latin characters in Japanese words, "T\begin{CJK}{UTF8}{min}シャツ\end{CJK}" (T-shirt) for instance, are also normalized similarly. After that slight pre-processing step, we merge all the QAs in all three languages into one dataset.

To prepare for the VLSP 2022 - EVJVQA Challenge, we split the dataset into training set, public test set, and private test set. We provide 3,763 images for the training set, while each test set comprises 558 images. Since no image is placed in multiple sets at once, we ensure that all QAs for a particular image only appear in the set to which that image belongs. The corresponding number of QAs in each language for each of the sets is shown in Table \ref{tab:num-qa}.

\begin{table}[ht]
\centering
\resizebox{0.5\textwidth}{!}{
\begin{tabular}{lcccc}
    \hline
                    & \textbf{Training} & \textbf{Public test} & \textbf{Private test} & \textbf{Total} \\ \hline
\textbf{Vietnamese} & 8,334             & 1,685                & 1,670                 & 11,689         \\
\textbf{English}    & 7,189             & 1,679                & 1,671                 & 10,539         \\
\textbf{Japanese}   & 8,262             & 1,651                & 1,649                 & 11,562         \\
\textbf{Total}      & 23,785            & 5,015                & 4,990                 & 33,790        
\\ \hline
\end{tabular}}
\caption{Number of QAs in each language in our UIT-EVJVQA dataset.}
\label{tab:num-qa}
\end{table}

\subsection{Statistics}

As our dataset was constructed from three languages: Vietnamese, English, and Japanese, we conducted statistics to deeply observe the characteristic of each language as well as gain clear insights into the three languages.

\begin{table}[ht]
    \centering
    \resizebox{0.5\textwidth}{!}{
    \begin{tabular}{lcccccc}
    \hline
    \multirow{2}{*}{}   & \multicolumn{3}{c}{\textbf{Question}}        & \multicolumn{3}{c}{\textbf{Answer}}           
    \\ \cline{2-7} 
                        & \textbf{Max.} & \textbf{Min.} & \textbf{Avg.} & \textbf{Max.} & \textbf{Min.} & \textbf{Avg.} \\ \hline
    \textbf{Vietnamese} & 22            & 3             & 8.7           & 32            & 1             & 7.2           \\
    \textbf{English}    & 26            & 3             & 8.6           & 23            & 1             & 5.0           \\
    \textbf{Japanese}   & 45            & 4             & 13.3          & 23            & 1             & 5.9           \\ \hline
    \end{tabular}}
    \caption{Statistic of question and answer in the UIT-EVJVQA dataset.}
    \label{tab:question_answer}
\end{table}

To conduct statistics on Vietnamese, we use the word segmentation method from the VnCoreNLP \cite{vu-etal-2018-vncorenlp} as in Vietnamese, a word may have more than one token (for instance, "cửa hàng tạp hóa" is formed from two Vietnamese words "cửa hàng" and "tạp hóa", which is in turn formed from more than one token). For English QAs, we achieved tokens by splitting sentences using space. For Japanese QAs, like Vietnamese, Japanese uses hieroglyphs to form their word, hence each Japanese word may have more than one hieroglyph. We use the janome\footnote{https://github.com/mocobeta/janome} library to perform word segmentation on Japanese QAs.

\begin{figure*}[ht]
    \centering
    \begin{subfigure}{0.315\textwidth}
        \includegraphics[width=\textwidth]{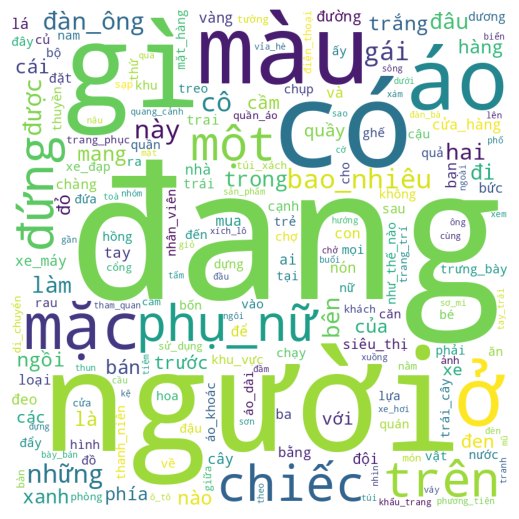}
        \caption{Vietnamese}
        \label{fig:vi_wordcloud}
    \end{subfigure}
    \begin{subfigure}{0.315\textwidth}
        \includegraphics[width=\textwidth]{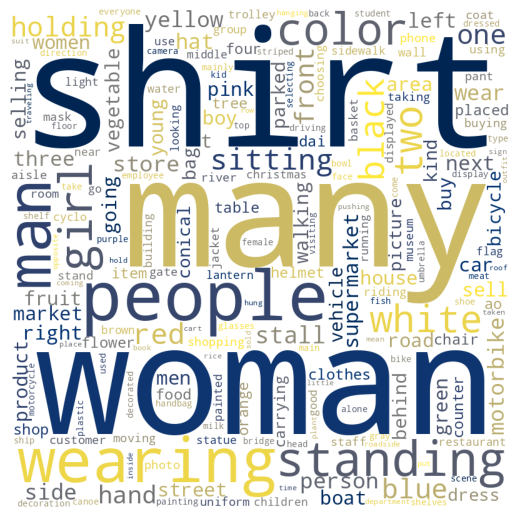}
        \caption{English}
        \label{fig:en_wordcloud}
    \end{subfigure}
    \begin{subfigure}{0.32\textwidth}
        \includegraphics[width=\textwidth]{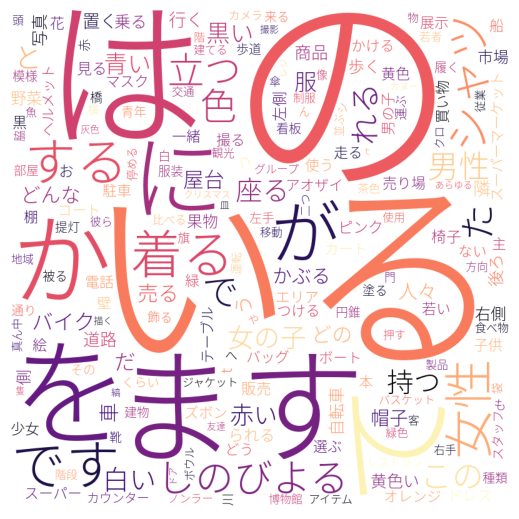}
        \caption{Japanese}
        \label{fig:jp_wordcloud}
    \end{subfigure}
    \caption{Word cloud of tokens in three language partitions of the UIT-EVJVQA dataset.}
\end{figure*}

Table \ref{tab:question_answer} indicate, with the same meaning, Japanese in general use more words to describe than Vietnamese and English. This implies another challenge for VQA method when tackling Japanese text beside the complexity of multilingualism in our dataset. Moreover, Vietnamese and English have the same distribution of length in terms of questions (according to Table \ref{tab:question_answer}), while English has shorter answers compared with those in Vietnamese. 

\begin{figure}[ht]
    \centering
    \begin{subfigure}[b]{0.5\textwidth}
         \centering
         \includegraphics[width=\textwidth]{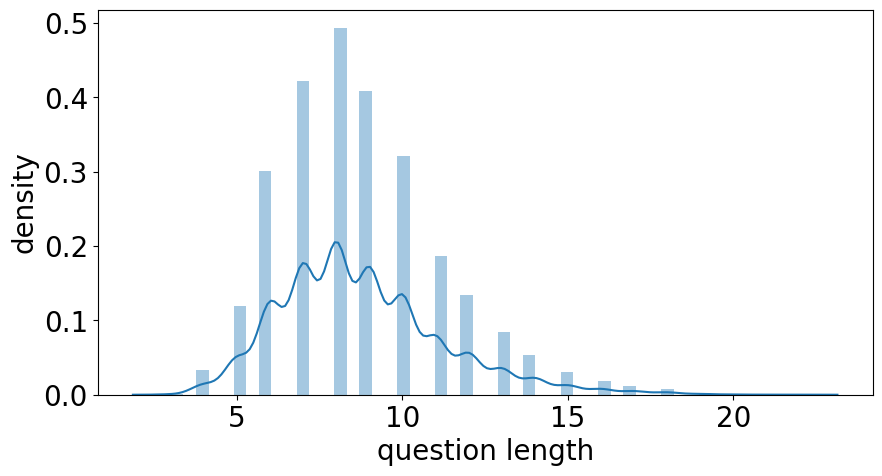}
    \end{subfigure}
    \begin{subfigure}[b]{0.5\textwidth}
         \centering
         \includegraphics[width=\textwidth]{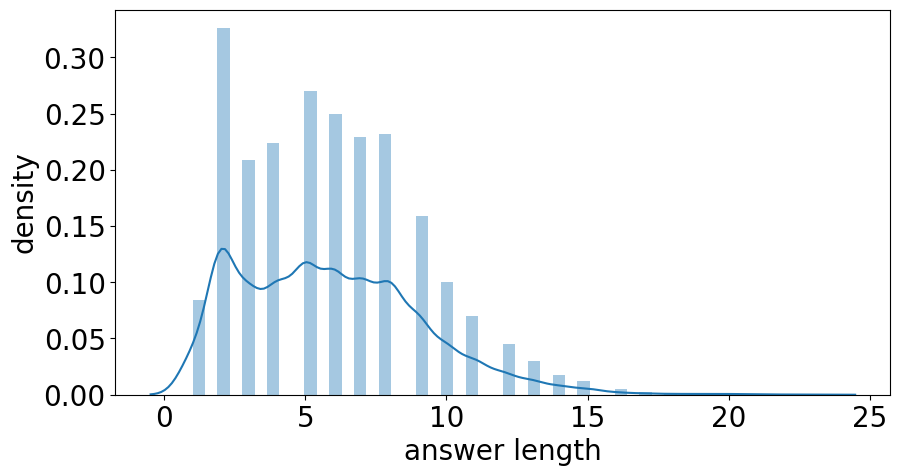}
    \end{subfigure}
    \caption{Statistics of the length of QAs in Vietnamese partition of the UIT-EVJVQA dataset.}
    \label{fig:vi_statistic}
\end{figure}

\begin{figure}[ht]
    \centering
    \begin{subfigure}[b]{0.5\textwidth}
         \centering
         \includegraphics[width=\textwidth]{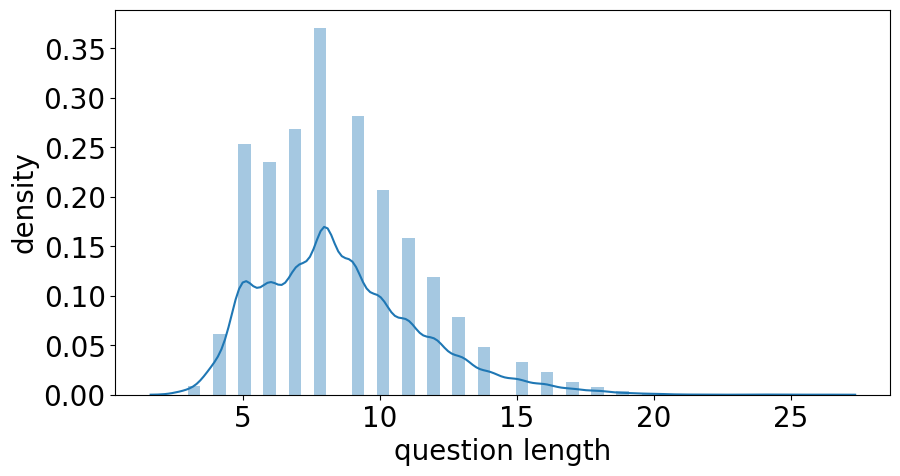}
    \end{subfigure}
    \begin{subfigure}[b]{0.5\textwidth}
         \centering
         \includegraphics[width=\textwidth]{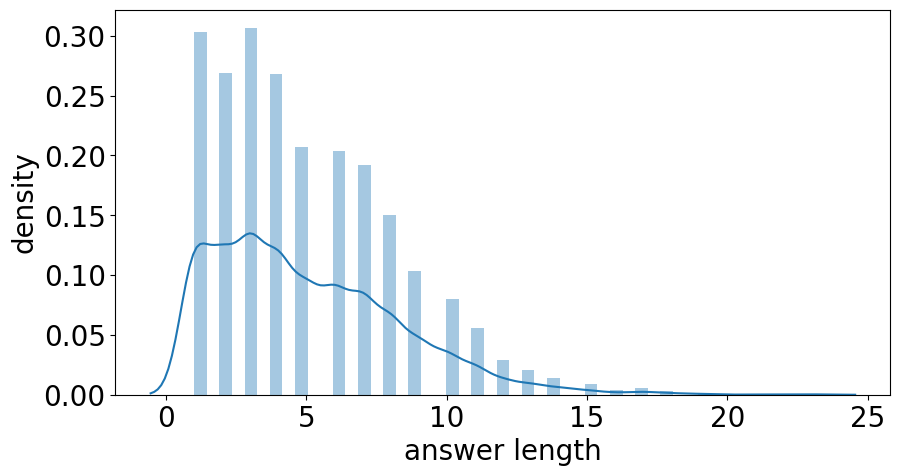}
    \end{subfigure}
    \caption{Statistics of the length of QAs in English partition of the UIT-EVJVQA dataset.}
    \label{fig:en_statistic}
\end{figure}

\begin{figure}[ht]
    \centering
    \begin{subfigure}[b]{0.5\textwidth}
         \centering
         \includegraphics[width=\textwidth]{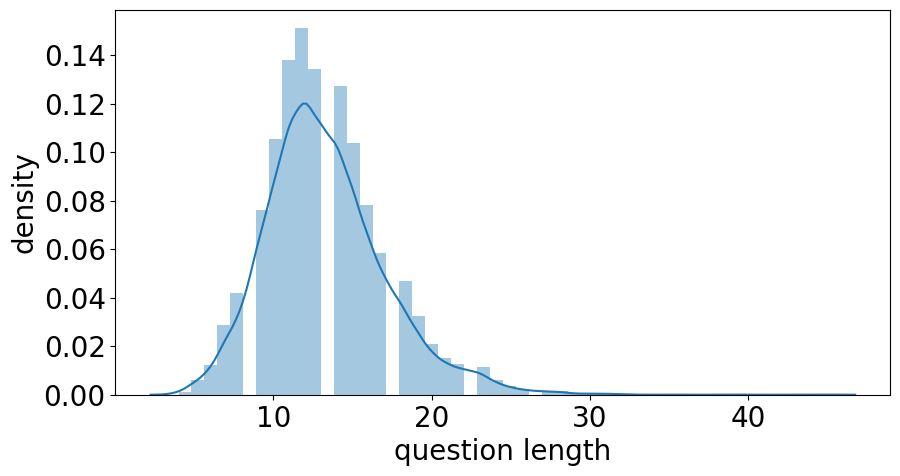}
    \end{subfigure}
    \begin{subfigure}[b]{0.5\textwidth}
         \centering
         \includegraphics[width=\textwidth]{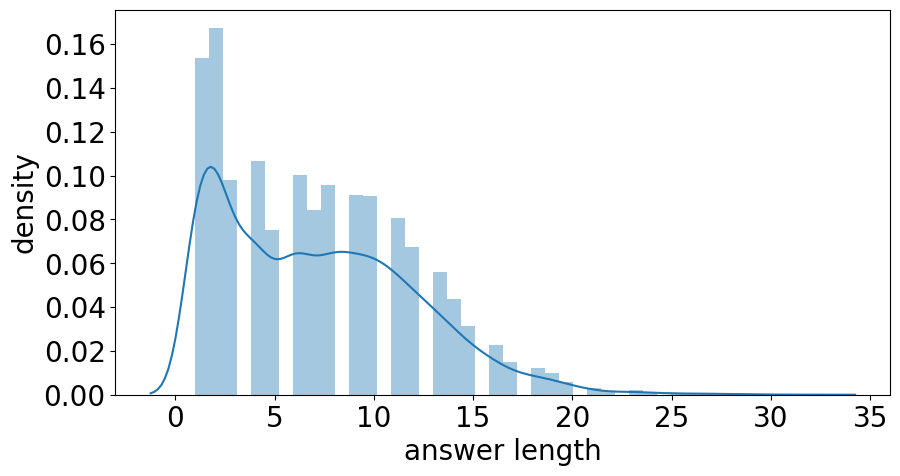}
    \end{subfigure}
    \caption{Statistics of the length of QAs in Japanese partition of the UIT-EVJVQA dataset.}
    \label{fig:jp_statistic}
\end{figure}

Answers of the three languages share the same characteristic where the most appearance of length is two. This indicates humans while giving answers, prefer saying in short statements, and this behavior leads to the classification approach on the VQA dataset \cite{teney2018tips}. However, such short answers are not always the case in our dataset as the context of images taken in Vietnam is complicated because of the crowded scenes, traffic jams, or street stalls, and short answers are not enough to answer questions enquiring about complex scenes. Moreover, we aim to emphasize the language aspect of the VQA task, which means we want to guide the community to research and propose a system that can give answers flexibly and naturally as a human does, not the way of "selecting" answers from a defined set as most approaches in the VQA dataset \cite{antol2015vqa,teney2018tips}. To this end, the answers given in our dataset are diverse in length and complicated in terms of level (word, phrase, or sentence level). Interestingly, while annotating answers, we found that giving a phrase or sentence as an answer is more fluent and human-like than giving only words or phrases as in the VQA dataset of \cite{antol2015vqa}.

Another factor to consider while annotating the UIT-EVJVQA dataset is the language prior phenomenon \cite{goyal2017making}. This is the phenomenon where the VQA methods try to learn patterns between questions and answers, such as questions starting with "how many" usually go with the answer "two". As analyzed in \cite{goyal2017making}, the language priors in VQA dataset is the result of the classification approach proposed for VQA task \cite{teney2018tips} and cause the model to learn the way of recognizing answer based on the question rather than the way of how to make use of the image to answer the given question. Therefore while constructing the guideline to annotate the UIT-EVJVQA dataset, we propose to give answers using words, phrases, or sentences. In this way, we can first eliminate the traditional classification approach proposed for the VQA task in English as well as avoid the language priors in our dataset.

%% file: Sections/5-Experiments.tex
\section{Systems and Results}

The aim of the challenge is to evaluate the quality
of the teams' approaches to multilingual visual question-answering systems.

\subsection{Baseline System}
\begin{figure}[ht]
    \centering
    \includegraphics[width=0.47\textwidth]{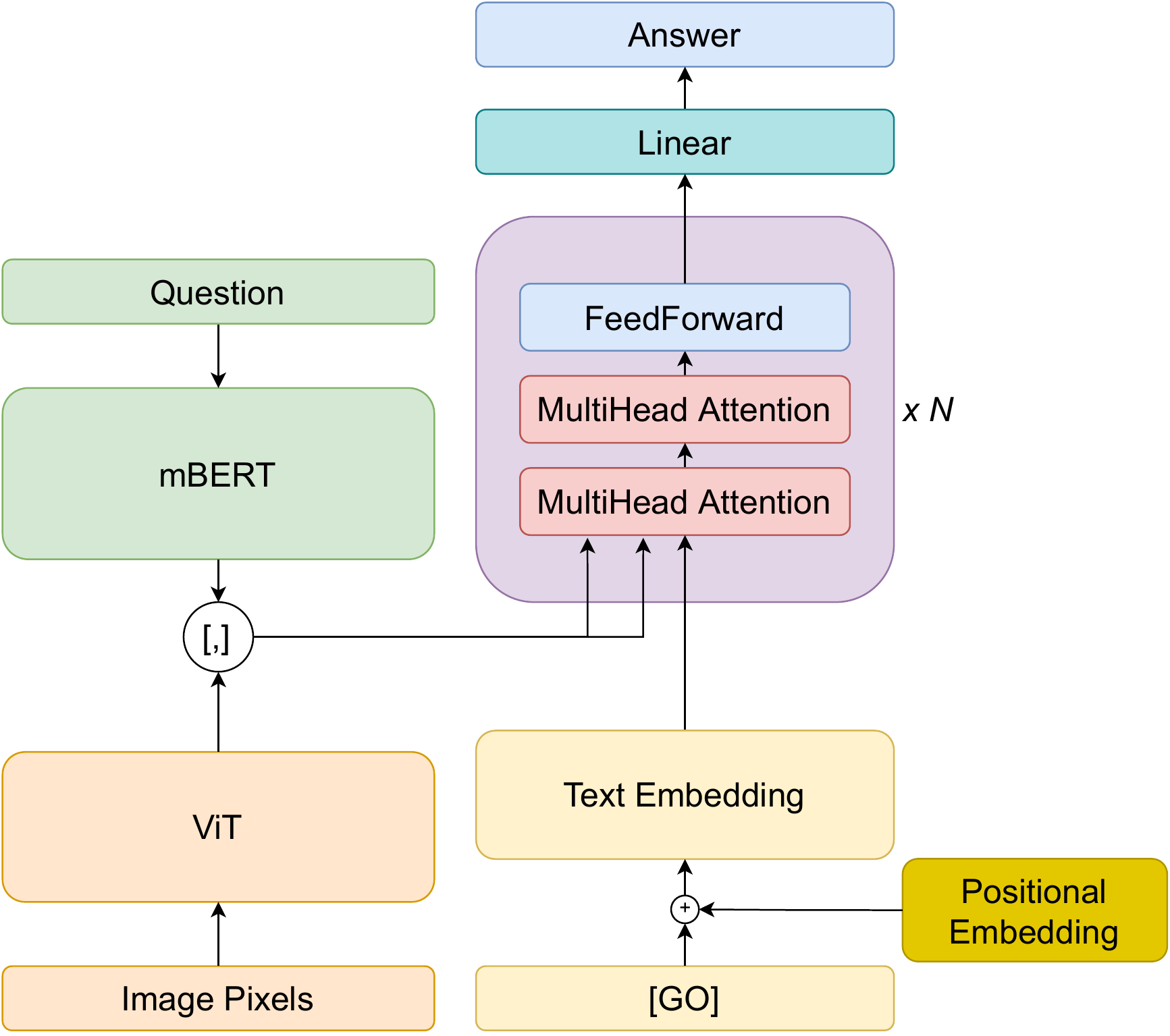}
    \caption{The baseline model architecture at the VLSP2022-EVJVQA Challenge.}
    \label{fig:baseline}
\end{figure}

Following Changpinyo et al. \cite{https://doi.org/10.48550/arxiv.2209.05401}, we adopt transfer learning based on Vision Transformer (ViT) \cite{dosovitskiy2020image} and mBERT \cite{devlin2018bert} for our baseline system. This work uses the training set for fine-tuning the pre-trained mBERT \cite{devlin2018bert} model before generating answers. In addition, we use the ViT \cite{dosovitskiy2020image} to extract the visual features of images (Figure \ref{fig:baseline}). Pre-trained ViT and mBERT models are initialized from HuggingFace checkpoints. We trained the baseline with a batch size of 64 and adapted the learning rate scheduler from Vaswani et al. \cite{vaswani2017attention} to reduce the learning rate gradually after a number of iterations. The training process was interrupted automatically if the evaluation scores did not increase after 5 epochs.

\subsection{Challenge Submission}

The competition was hosted offline where the organizers provide the training set and public test set to the participant teams to evaluate and fine-tune their methods. When the competition came into the private test phase, the submission policy allows each participant team to submit up to 3 \textit{different} methods during a submission time lasting 3 days. After the three-day private test phase, we evaluated their submitted results and obtain their F1 score as well as avg. BLEU score on the private test set. The final score of each participant team is the highest score among their submitted models on the private test set and their rank was indicated based on the F1 score ((BLUE as a secondary score when there is a tie).

\subsection{Experimental Results}

\begin{table*}[ht]
\centering
\resizebox{1\textwidth}{!}{
\begin{tabular}{llccllllll}
\hline
\multicolumn{1}{c}{\multirow{2}{*}{No.}} & \multicolumn{1}{c}{\multirow{2}{*}{Team name}} & \multicolumn{1}{l}{\multirow{2}{*}{Model type}} & \multicolumn{3}{c}{\multirow{2}{*}{Models}} & \multicolumn{2}{c}{Public Test} & \multicolumn{2}{c}{Private Test} \\ \cline{7-10} 
\multicolumn{1}{c}{} & \multicolumn{1}{c}{} & \multicolumn{1}{l}{} & \multicolumn{3}{c}{} & \multicolumn{1}{c}{F1} & \multicolumn{1}{c}{BLEU} & \multicolumn{1}{c}{F1} & \multicolumn{1}{c}{BLEU} \\ \hline
\textbf{1} & \textbf{CIST AI} & Single & \multicolumn{3}{c}{ViT + mT5} & 0.3491 & 0.2508 & \textbf{0.4392} & \textbf{0.4009} \\
\textbf{2} & \textbf{OhYeah} & Single & \multicolumn{3}{c}{ViT + mT5} & \textbf{0.5755} & \textbf{0.4866} & 0.4349 & 0.3868 \\
\textbf{3} & \textbf{DS-STBFL} & Ensemble & \multicolumn{3}{c}{CNN-Seq2Seq + ViT + OFA} & 0.3390 & 0.2156 & 0.4210 & 0.3482 \\
4 & FCoin & Single & \multicolumn{3}{c}{ViT + mBERT} & 0.3355 & 0.2437 & 0.4103 & 0.3549 \\
5 & VL-UIT & Single & \multicolumn{3}{c}{\begin{tabular}[c]{@{}c@{}}BEiT + CLIP + Detectron-2\\ + mBERT + BM25 + FastText\end{tabular}} & 0.3053 & 0.1878 & 0.3663 & 0.2743 \\
6 & BDboi & Ensemble & \multicolumn{3}{c}{\begin{tabular}[c]{@{}c@{}}ViT + BEiT + SwinTransformer\\ + CLIP + OFA + BLIP\end{tabular}} & 0.3023 & 0.2183 & 0.3164 & 0.2649 \\
7 & UIT-squad & Ensemble & \multicolumn{3}{c}{VinVL+mBERT} & 0.3224 & 0.2238 & 0.3024 & 0.1667 \\
8 & VC-Internship & Single & \multicolumn{3}{c}{ResNet-152 + OFA} & 0.3017 & 0.1639 & 0.3007 & 0.1337 \\
\textit{9} & \textit{Baseline} & Single & \multicolumn{3}{c}{ViT + mBERT} & \textit{0.2924} & \textit{0.2183} & \textit{0.3346} & \textit{0.2275} \\ \hline
\end{tabular}
}
\caption{Final results of submitted methods.}
\label{tab:my-table}
\end{table*}

There are 8 participant teams submitted in total, and 5 of them have F1 scores higher than the baseline system.

%% file: Sections/6-Result-Analysis.tex
\section{Result Analysis} \label{section-6}

We mainly analyze the results of methods coming from participating teams based on the length of questions and answers in each language. To alleviate the result analysis, we define the four different ranges for the length of questions and answers. In particular, we define the \textit{short questions} (respective, \textit{short answer}) are questions whose length is shorter or equal to 5 tokens, \textit{medium questions} (respective, \textit{medium answer}) questions whose length is between 6 to 10 tokens, \textit{long questions} (respective, \textit{long answer}) are questions whose length is between 11 to 15 tokens, and finally, \textit{very long questions} (respective, \textit{very long answer}) are questions whose length is greater than 15 tokens. Tokens of questions and answers are defined in the same manner when we do statistics in Section 4.4. We sequentially report the results of the top-5 models in terms of scores and visualization in quantitative analysis and qualitative analysis, respectively, to support our statements as well as indicate what the research community should pay attention to when constructing the novel open-ended VQA dataset.

\begin{table*}[ht]
    \centering
    \resizebox{\textwidth}{!}{
    \begin{tabular}{cccccccccccc}
    \hline
    \multicolumn{2}{c}{\multirow{2}{*}{}} & \multicolumn{2}{c}{\textbf{Top 1}} & \multicolumn{2}{c}{\textbf{Top 2}} & \multicolumn{2}{c}{\textbf{Top 3}} & \multicolumn{2}{c}{\textbf{Top 4}} & \multicolumn{2}{c}{\textbf{Top 5}} \\ \cline{3-12}
\multicolumn{2}{c}{} & \textbf{F1}     & \textbf{BLEU}     & \textbf{F1}     & \textbf{BLEU}     & \textbf{F1}     & \textbf{BLEU}     & \textbf{F1}     & \textbf{BLEU}     & \textbf{F1}     & \textbf{BLEU}    \\ \hline
\multirow{4}{*}{\rotatebox{90}{\textbf{Question}}} & S                              & 0.3900          & 0.1867            & 0.5000          & 0.1250            & 0.4261          & 0.2022            & 0.4308          & 0.2184            & 0.3728          & 0.1535           \\
& M                              & 0.3935          & 0.2218            & 0.4922          & 0.2961            & 0.3833          & 0.1889            & 0.3730          & 0.1909            & 0.3222          & 0.1329           \\
& L                              & 0.3815          & 0.2044            & 0.4470          & 0.2832            & 0.3475          & 0.1628            & 0.3248          & 0.1549            & 0.2982          & 0.1093           \\
& XL                             & 0.3681          & 0.2019            & 0.4368          & 0.2838            & 0.3059          & 0.1518            & 0.3137          & 0.1253            & 0.3030          & 0.1288           \\ \hline
\multirow{4}{*}{\rotatebox{90}{\textbf{Answer}}} & S                              & 0.3000          & 0.1308            & 0.2429          & 0.0975            & 0.2999          & 0.1209            & 0.3004          & 0.1282            & 0.2498          & 0.0912           \\
& M                              & 0.4751          & 0.2912            & 0.3478          & 0.1994            & 0.4603          & 0.2530            & 0.4502          & 0.2553            & 0.4013          & 0.1737           \\
& L                              & 0.4742          & 0.2948            & 0.5063          & 0.3522            & 0.4473          & 0.2237            & 0.3837          & 0.1897            & 0.3642          & 0.1517           \\
& XL                             & 0.4456          & 0.1580            & 0.5709          & 0.4029            & 0.3882          & 0.1049            & 0.3798          & 0.1175            & 0.4219          & 0.1392          \\ \hline
\end{tabular}}
\caption{Results of the top-5 methods in English part on the private test set. S stands for short, M stands for medium, L stands for long, and XL stands for very long.}
\label{top_5_en}
\end{table*}

From Table \ref{top_5_en}, we have the top-5 models share the same characteristic on the English part of the UIT-EVJVQA dataset. Particularly, when we observe the results based on the question length, we can see that all submitted models have the same behavior when they give a higher performance on short questions and medium questions, while they yield a few drawbacks on long questions and very long questions. Turn the attention to the last four rows, which are the results based on answer length. We have different behavior. The top-5 models give better results on medium and long answers while yielding worse results on short and very long answers. Come from Figure \ref{fig:en_statistic}, we can see that most English answers are short answers, which means models have more short answers to learning hence logically, they should have better performance on short answers rather than medium answers and long answers. To answer this weird insight, we showed all answers given by the top-5 models, and we found out that all top-5 models tend to give medium answers, even for questions having short answers. The most exciting thing here is how models give medium or lengthy answers: they repeat some tokens from the questions, and this is the primary way our annotators give medium and lengthy answers to questions. For questions with medium or long answers, answers from top-5 models mostly repeat some tokens from questions, and the gold medium or gold answers also repeat some tokens from questions. Thanks to these matched tokens with questions, F1 scores, and avg. BLEU scores of the answers from top-5 models are pretty high, while the crucial tokens, which determine whether or not the information in these answers is correct or not, usually have a length of 2 or 3 tokens; in case of totally wrong, they still do not affect the overall scores significantly. To gain a better understanding, we show some samples in the Appendix for an intuitive explanation.

\begin{table*}[ht]
\centering
\resizebox{\textwidth}{!}{
\begin{tabular}{cccccccccccc}
\hline
\multicolumn{2}{c}{\multirow{2}{*}{}} & \multicolumn{2}{c}{\textbf{Top 1}} & \multicolumn{2}{c}{\textbf{Top 2}} & \multicolumn{2}{c}{\textbf{Top 3}} & \multicolumn{2}{c}{\textbf{Top 4}} & \multicolumn{2}{c}{\textbf{Top 5}} \\ \cline{3-12}
\multicolumn{2}{c}{\textbf{}} & \textbf{F1}     & \textbf{BLEU}     & \textbf{F1}     & \textbf{BLEU}     & \textbf{F1}     & \textbf{BLEU}     & \textbf{F1}     & \textbf{BLEU}     & \textbf{F1}     & \textbf{BLEU}    \\ \hline
\multirow{4}{*}{\rotatebox{90}{\textbf{Question}}} & S                              & 0.4143          & 0.2088            & 0.4370          & 0.2412            & 0.4582          & 0.2482            & 0.4894          & 0.2768            & 0.4116          & 0.1909           \\
& M                              & 0.4997          & 0.3259            & 0.4976          & 0.3150            & 0.5008          & 0.3111            & 0.4879          & 0.3016            & 0.4347          & 0.2271           \\
& L                              & 0.4797          & 0.2954            & 0.4582          & 0.2582            & 0.4600          & 0.2687            & 0.4423          & 0.2545            & 0.4014          & 0.1835           \\
& XL                             & 0.4197          & 0.2446            & 0.4349          & 0.2083            & 0.4430          & 0.2359            & 0.4121          & 0.2234            & 0.3154          & 0.1447           \\ \hline
\multirow{4}{*}{\rotatebox{90}{\textbf{Answer}}} & S                              & 0.3621          & 0.1659            & 0.3719          & 0.1700            & 0.3692          & 0.1666            & 0.3576          & 0.1656            & 0.3493          & 0.1481           \\
& M                              & 0.5564          & 0.3874            & 0.5507          & 0.3743            & 0.5467          & 0.3626            & 0.5433          & 0.3609            & 0.4705          & 0.2604           \\
& L                              & 0.5326          & 0.3623            & 0.4859          & 0.2849            & 0.5141          & 0.3244            & 0.4802          & 0.2851            & 0.4007          & 0.1760           \\
& XL                             & 0.4898          & 0.3172            & 0.4627          & 0.1959            & 0.5328          & 0.3253            & 0.4142          & 0.2243            & 0.3430          & 0.1278          \\ \hline
\end{tabular}}
\caption{Results of the top-5 methods in Vietnamese part on the private test set. S stands for short, M stands for medium, L stands for long, and XL stands for very long.}
\label{top_5_vi}
\end{table*}

Coming into the Vietnamese part of the UIT-EVJVQA dataset results, we have quite different behavior where most models perform better on medium and long questions. When we observe the length distribution of questions in Figure \ref{fig:vi_statistic}, most questions have a length range of around eight tokens, or most have medium length. While in English, most questions fell in the range of 5 tokens and seven tokens. Therefore the top-5 models have a good performance on short and medium questions in English, while they achieved better performance on medium and long answers in Vietnamese. For Vietnamese answers, as indicated in Table \ref{top_5_vi}, top-5 models share the same performance as the English answers: they perform better on medium and long answers than on short and very long answers. We also provided some samples in Appendix to better demonstrate the method with English answers.

\begin{table*}[ht]
\centering
\resizebox{\textwidth}{!}{
\begin{tabular}{cccccccccccc}
\hline
\multicolumn{2}{c}{\multirow{2}{*}{}} & \multicolumn{2}{c}{\textbf{Top 1}} & \multicolumn{2}{c}{\textbf{Top 2}} & \multicolumn{2}{c}{\textbf{Top 3}} & \multicolumn{2}{c}{\textbf{Top 4}} & \multicolumn{2}{c}{\textbf{Top 5}} \\ \cline{3-12}
\multicolumn{2}{c}{\textbf{}} & \textbf{F1}     & \textbf{BLEU}     & \textbf{F1}     & \textbf{BLEU}     & \textbf{F1}     & \textbf{BLEU}     & \textbf{F1}     & \textbf{BLEU}     & \textbf{F1}     & \textbf{BLEU}    \\ \hline
\multirow{4}{*}{\rotatebox{90}{\textbf{Question}}} & S                              & 0.7500          & 0.3125            & 0.4060          & 0.1917            & 0.6667          & 0.1850            & 0.4444          & 0.1250            & 0.3333          & 0.0460           \\
& M                              & 0.4494          & 0.2004            & 0.3905          & 0.2164            & 0.4634          & 0.2984            & 0.4841          & 0.3118            & 0.4485          & 0.2447           \\
& L                              & 0.4489          & 0.2841            & 0.3679          & 0.1974            & 0.4395          & 0.2720            & 0.4150          & 0.2741            & 0.4058          & 0.2333           \\
& XL                             & 0.4399          & 0.2915            & 0.3456          & 0.1863            & 0.3896          & 0.2333            & 0.3834          & 0.2453            & 0.3452          & 0.1815           \\ \hline
\multirow{4}{*}{\rotatebox{90}{\textbf{Answer}}} & S                              & 0.1983          & 0.075             & 0.3101          & 0.1336            & 0.2057          & 0.0713            & 0.1809          & 0.0714            & 0.1983          & 0.0750           \\
& M                              & 0.3503          & 0.1942            & 0.4654          & 0.2822            & 0.3236          & 0.1750            & 0.3443          & 0.2095            & 0.3503          & 0.1942           \\
& L                              & 0.5242          & 0.3667            & 0.4436          & 0.2731            & 0.4734          & 0.3156            & 0.4609          & 0.3230            & 0.5242          & 0.3667           \\
& XL                             & 0.5911          & 0.4245            & 0.3684          & 0.1586            & 0.5131          & 0.3321            & 0.4983          & 0.3358            & 0.5911          & 0.4245          \\ \hline
\end{tabular}}
\caption{Results of the top-5 methods in Japanese partition on the private test set in. S stands for short, M stands for medium, L stands for long, and XL stands for very long.}
\end{table*}
\label{top_5_jp}

On the Japanese part of the UIT-EVJVQA dataset, the top-5 models have the same color on the English part when they give better results on medium and long questions. But unlike their performance in Vietnamese and English, where they achieved better scores for medium and long answers, the top-5 models have results as increase as the length of answers. From Figure \ref{fig:jp_statistic}, although the occurrence of short answers is the highest, the cumulative of medium and long answers are higher than the cumulative of short answers, which indicates the top-5 models not only have more medium and lengthy answers to learn but also tend to give medium or long answers to have the optimal loss on the training set. We also provided some samples in the Appendix for better demonstration.

Apart from the previous discussion, we can conclude on the UIT-EVJVQA dataset, most deep learning models tend to give lengthy answers to given questions with images, and the way they give lengthy answers is by repeatedly using some tokens of questions as starting point of the answers, and the main wrong parts of these answers are at the tokens indicating vital information to answer the questions such as objects, colors or side (see Appendix for more examples). Hence, to better understand how wrong the top-5 model gives predictions, we conduct analyses focusing on the usage of side, object, and color words in each language.

\begin{figure*}[ht]
    \centering
    \begin{subfigure}{0.32\textwidth}
        \includegraphics[width=0.95\textwidth]{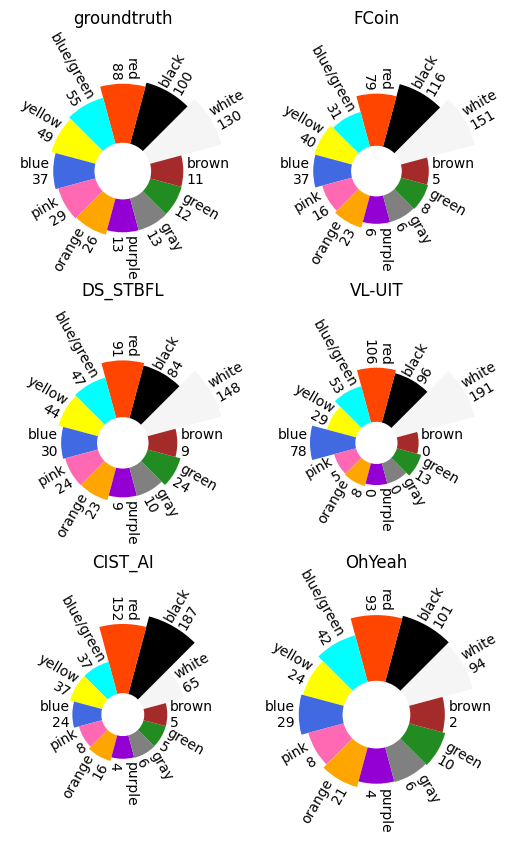}
        \caption{Vietnamese}
        \label{fig:vn_color}
    \end{subfigure}
    \begin{subfigure}{0.32\textwidth}
        \includegraphics[width=\textwidth]{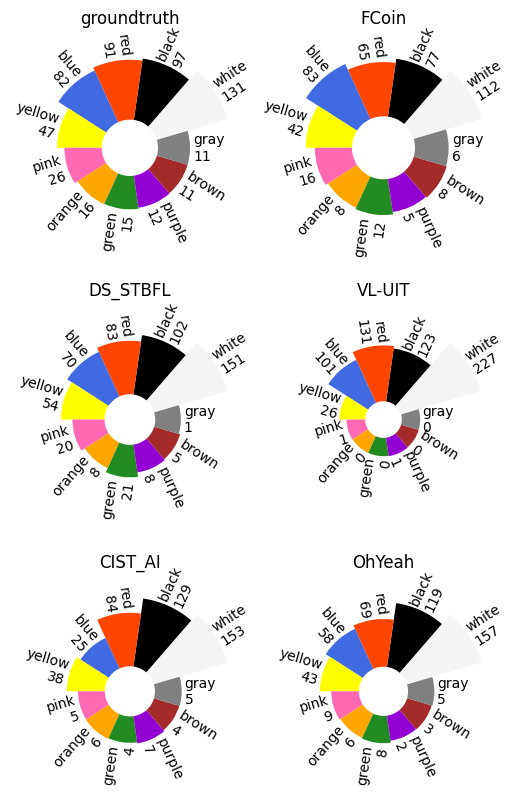}
        \caption{English}
        \label{fig:en_color}
    \end{subfigure}
    \begin{subfigure}{0.32\textwidth}
        \includegraphics[width=\textwidth]{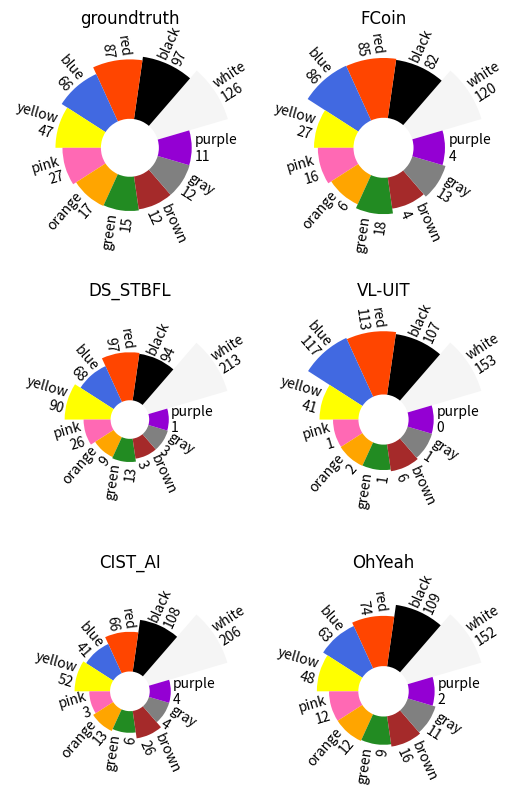}
        \caption{Japanese}
        \label{fig:jp_color}
    \end{subfigure}
    \caption{Color words distribution in ground truth and predictions of each team in three languages.}
\end{figure*}

\textbf{Side words}: One of the most confusing attributes while inferring the description of objects is the use of side or direction. This is the case where an object is observed on the left or right of another object or appears on the left or right side of the scene. In English, side words are the words "left" and "right" in our dataset. While in Japanese, they are "\begin{CJK}{UTF8}{min}左\end{CJK}" and "\begin{CJK}{UTF8}{min}右\end{CJK}". However, they are not simply "trái" and "phải" in Vietnamese, as "trái" may not often be used to convey a side. Therefore, we broaden the set of side words 
as "bên trái", "bên phải", "tay trái" (left-hand side) and "tay phải" (right-hand side) according to the observation of the dataset. To measure how well each submitted model uses those side words in each language, we adopt the F1-score to calculate the proportion of match in terms of side words between the predictions and the gold answers. According to the results in Table \ref{tab:side-f1}, most models failed to indicate the side of objects while answering the questions.

\begin{table}[ht]
\centering
\begin{tabular}{llll}
\hline
\multicolumn{1}{c}{Team} & \multicolumn{1}{c}{Vietnamese} & \multicolumn{1}{c}{English} & \multicolumn{1}{c}{Japanese} \\ \hline
CIST AI                 & \textbf{0.4948}                & \textbf{0.3889}             & 0.3922                       \\
OhYeah                   & 0.3814                         & 0.3235                      & \textbf{0.4085}              \\
DS-STBFL                & 0.4811                         & 0.3137                      & 0.3366                       \\
FCoin                    & 0.4021                         & 0.3039                      & 0.3595                       \\
VL-UIT                   & 0.2268                         & 0.2418                      & 0.1471                       \\ \hline
\end{tabular}
\caption{F1-score for side predictions of every team for each language}
\label{tab:side-f1}
\end{table}

\textbf{Color words}: As an attribute frequently appears in the QAs of our dataset, color is worth being observed as part of the inference performance of the models. Here we measure the F1-score of color word matches between predicted and gold answers for every team. The color words have been defined in the guideline in Section \ref{guideline}. However, there are some minor exceptions due to the crowdsourcing process.

\begin{table}[ht]
\centering
\begin{tabular}{llll}
\hline
\multicolumn{1}{c}{Team} & \multicolumn{1}{c}{Vietnamese} & \multicolumn{1}{c}{English} & \multicolumn{1}{c}{Japanese} \\ \hline
CIST AI                 & 0.3230                         & 0.3434                      & 0.3450                       \\
OhYeah                   & 0.2954                         & 0.3326                      & 0.3229                       \\
DS-STBFL                & \textbf{0.5161}                & \textbf{0.4933}             & \textbf{0.3934}              \\
FCoin                    & 0.3619                         & 0.3142                      & 0.3234                       \\
VL-UIT                   & 0.2716                         & 0.2420                      & 0.2355                       \\ \hline
\end{tabular}
\caption{F1-score for color predictions of every team for each language}
\label{tab:color-f1}
\end{table}

\begin{figure*}
    \centering
    \includegraphics[width=\textwidth]{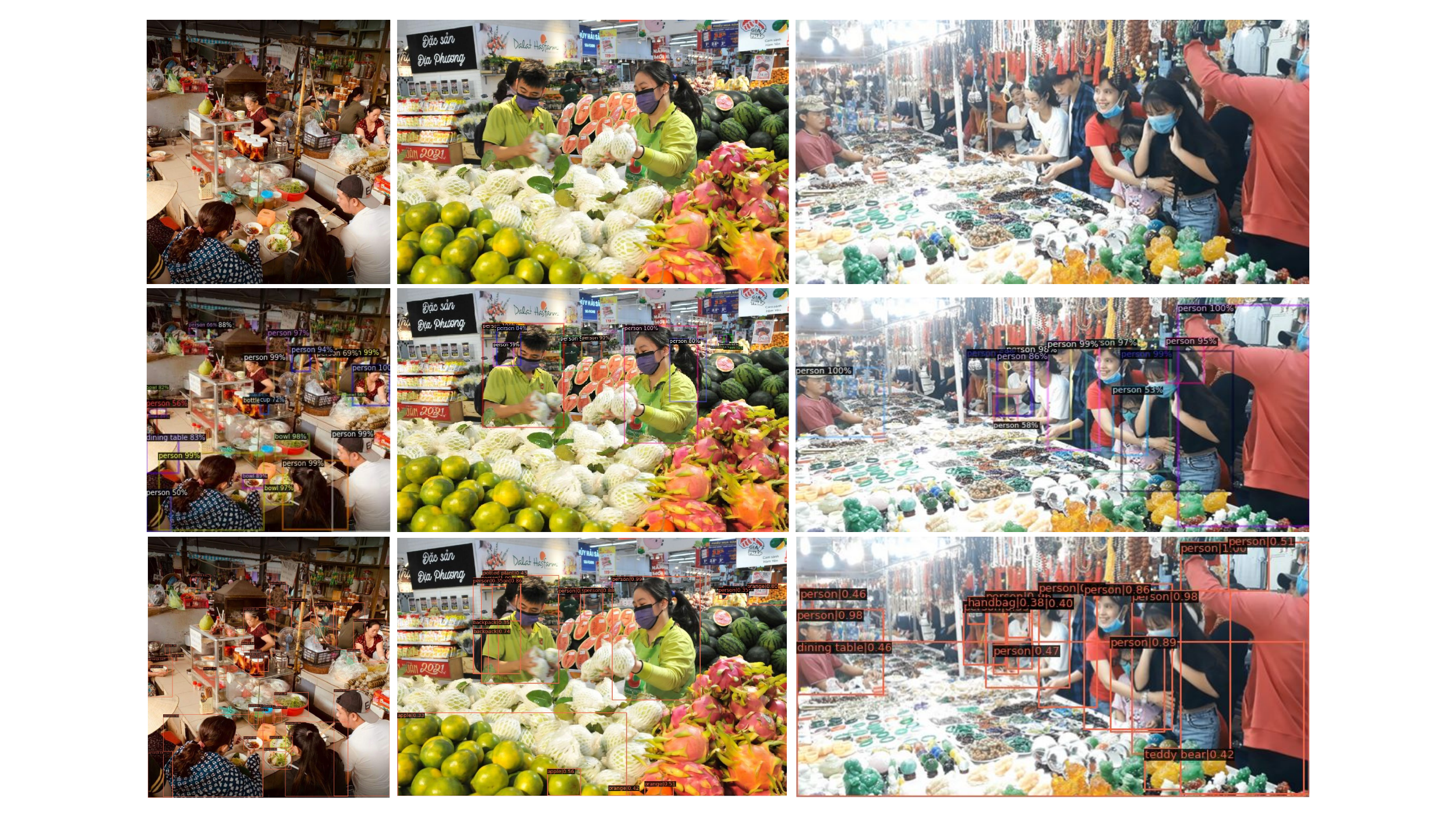}
    \caption{Some examples for the performance of pre-trained image models. From top to bottom: origin images, object detection results of Faster-RCNN with Resnet101 \cite{he2016deep} as backbone, and object detection results of Cascade RCNN \cite{cai2018cascade} with Swin Transformer \cite{liu2021swin} as the backbone (Zooming out x4 times for better illustration as well as clearly observing the detected labels and their confident scores of each model).}
    \label{fig:wrong_cases_image_models}
\end{figure*}

We also visualize the color word distribution in the prediction of every team for each language. It is worth noting that in the Vietnamese QAs, there is a substantial quantity of "xanh" color words which are unclear to be whether green or blue, but rather depend on the context of the visual scene. As we can see in Figure \ref{fig:vn_color}, Figure \ref{fig:en_color} and Figure \ref{fig:jp_color}, the ground truth color words are not evenly distributed because some colors such as white, black and red are used more frequently in the dataset, while the instances of brown, gray and purple are scant. This skewness is then emphasized through the overall inference of models. In Japanese, most submitted models clearly express bias as they intensively describe objects in white color, while less-appeared colors are poorly used. This kind of behavior is similar to the prior-language phenomenon pointed out in the previous work \cite{goyal2017making} as when being given questions about colors, top-5 models tend to use most-appearance colors despite the colors in images, and this phenomenon is also noticeable in other languages for some teams. For instance, the submitted model of team VL-UIT even ignores many of the less-appeared colors. As an attribute that is regularly used to describe an object, color also plays an important role in distinguishing between objects, which is an essential factor in visual question answering. Therefore, more efforts must be conducted to degrade the skewness in the color distribution in a way that helps the model infer better and more precisely describe objects out of images. We suggest future works should define a fixed range of colors and pay attention while asking and using colors to answer in order that we can ignore such prior-language phenomena on the VQA dataset.

\textbf{Objects}: To get the distribution of objects in the UIT-EVJVQA dataset, we used the POS method from the previous work \cite{vu-etal-2018-vncorenlp} and achieved objects by collecting tokens tagged as nouns. We used the same manner as investigating the behavior of top-5 models on color words but we can not find such language-prior phenomenon on objects. To find out the reason why most of the top-5 models failed in recognizing objects in images, we observed the results of pre-trained image models. However, most of the top-5 model's used pre-trained image models that output grid-based features \cite{jiang2020defense} such as ViT \cite{dosovitskiy2020image} or BEiT \cite{bao2021beit}, which means it is hard to visualize and interpret the results of those image models. Nevertheless, we can interpret results of other similar pre-trained image models such as Faster-RCNN \cite{ren2015faster} used ResNet101 \cite{he2016deep} as the backbone or Swin Transformer for object detection \cite{liu2021swin} as these models were pre-trained on the ImageNet dataset \cite{russakovsky2015imagenet} as well before being fine-tuned on various tasks, hence we used the outcomes of these two models on the training set of the UIT-EVJVQA dataset as an approximate way to investigate the reason of failure in recognizing an object in images of the top-5 models. As depicted in Figure \ref{fig:wrong_cases_image_models}, these two image models accurately detect objects that they were trained, but lots of common objects in Vietnam such as non la or fan (the first column of three images in Figure \ref{fig:wrong_cases_image_models} or some traditional products available in most of the markets in Vietnam (the last column of three images in Figure \ref{fig:wrong_cases_image_models}) are not detected, and some of the detected objects in Figure \ref{fig:wrong_cases_image_models} are not correct. This result 
implies the incorrect image understanding of pre-trained images models trained on images captured outside of Vietnam and indirectly affects the performance of VQA models (in case of using grid-based features, as region-based features \cite{jiang2020defense} were achieved from grid-based features, e.g. from the backbone of Faster-RCNN, hence the failures indicated from region-based features indicates as well the failure in grid-based features). In conclusion available pre-trained image models are not relevant to scenes captured in Vietnam and the Computer Vision (CV) community in Vietnam should research and develop a better appropriate pre-trained image model, especially for images taken in Vietnam so that we can effectively tackle recent trending tasks where multi modeling task such as VQA is one of them.

%% file: Sections/7-Conclusion.tex
\section{Conclusion and Future Work}

The VLSP2022-EVJVQA Challenge on multilingual image-based question answering has been organized at the VLSP 2022. Even though 57 teams had legally signed up to get the training dataset, only eight teams submitted their results. Because several teams enrolled for many challenges at the VLSP 2022, the other teams may not have enough time to explore VQA models. The highest performances are 0.4392 in F1-score and 0.4009 in BLUE on the private test set. The multilingual VQA systems proposed by the top 2 teams use ViT for the pre-trained vision model and mT5 for the pre-trained language model. EVJVQA is a challenging dataset including the training set, the development set (public test set), and the test set (private test set) that motivates NLP and CV researchers to further explore the multilingual models or systems in visual question answering.

To increase performance in multilingual visual question answering, we intend to increase the amount and quality of annotated questions in the future. In addition, we also make human-adversarial questions based on findings proposed by the research work \cite{sheng2021human}.

%% file: Sections/Appendix.tex
\appendix

\section{Appendix} \label{appendix}
In this section we provide extensively samples from top-5 models on the three languages for better demonstrating out observation in Section \ref{section-6}.

\begin{figure*}
    \centering
    \includegraphics[width=\textwidth]{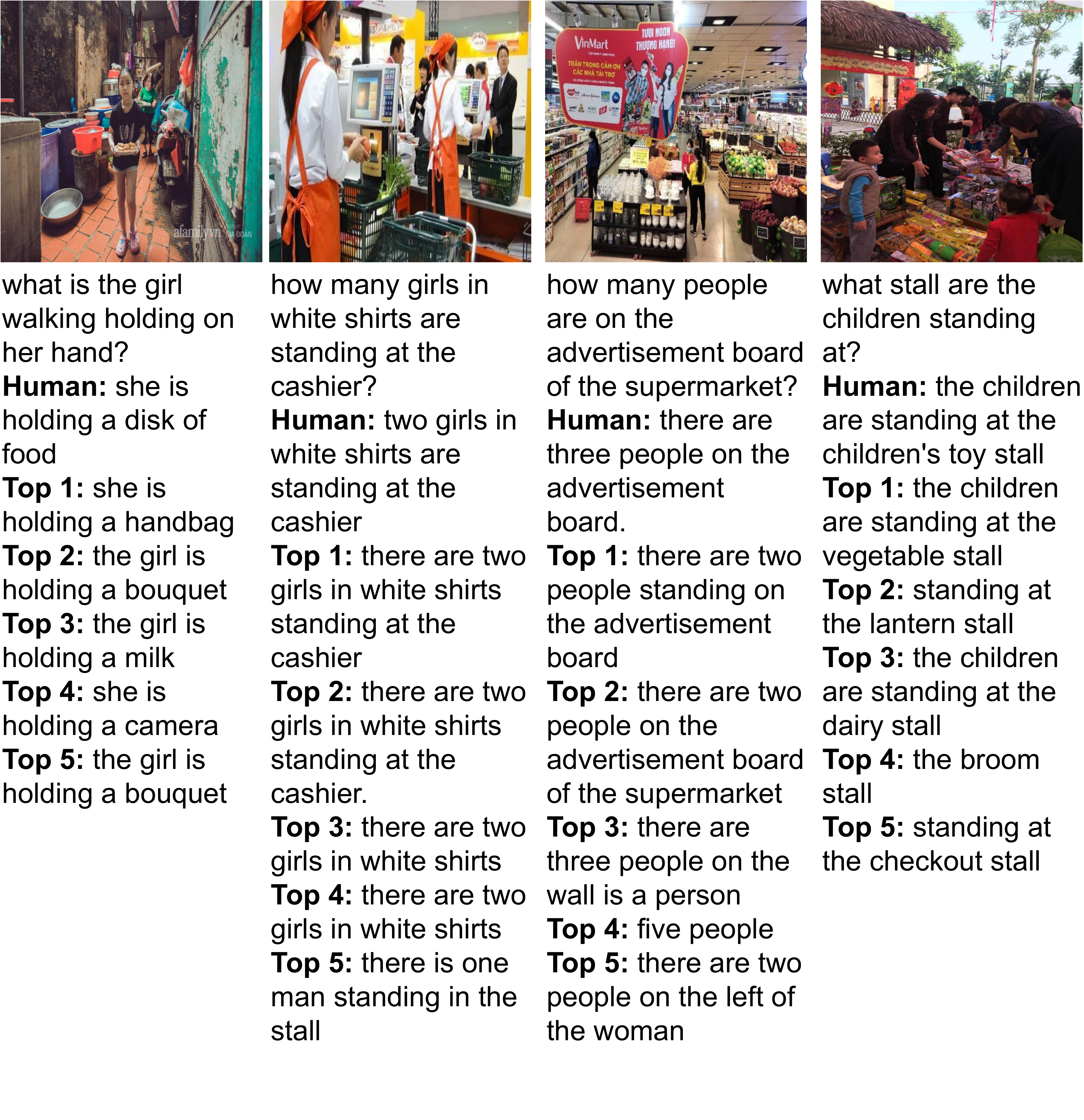}
    \caption{Results of top-5 models on medium answers in English of the UIT-EVJVQA dataset}
    \label{fig:results_en_medium}
\end{figure*}

\begin{figure*}
    \centering
        \includegraphics[width=\textwidth]{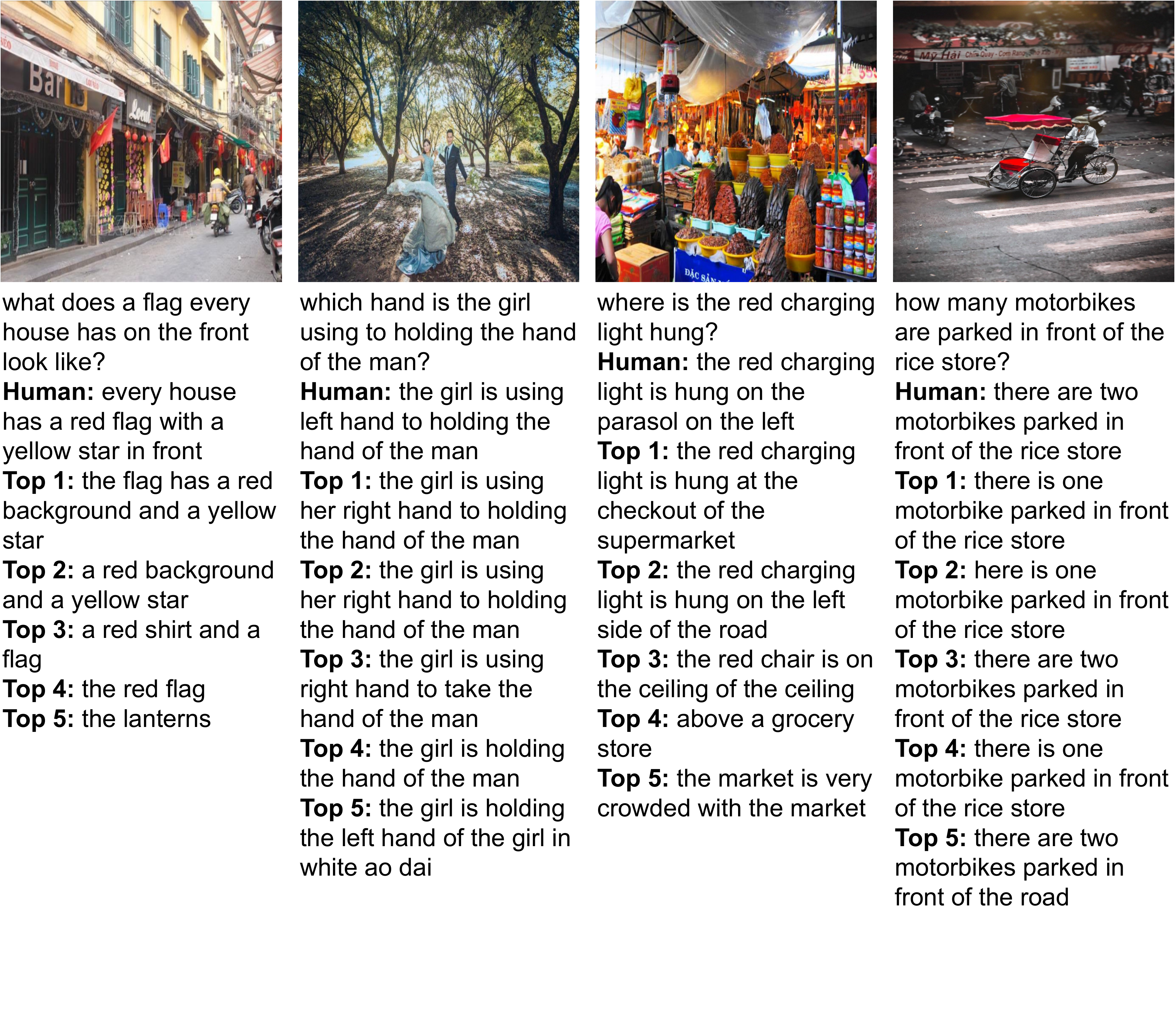}
    \caption{Results of top-5 models on long answers in English of the UIT-EVJVQA dataset}
    \label{fig:results_en_long}
\end{figure*}

\begin{figure*}
    \centering
    \includegraphics[width=\textwidth]{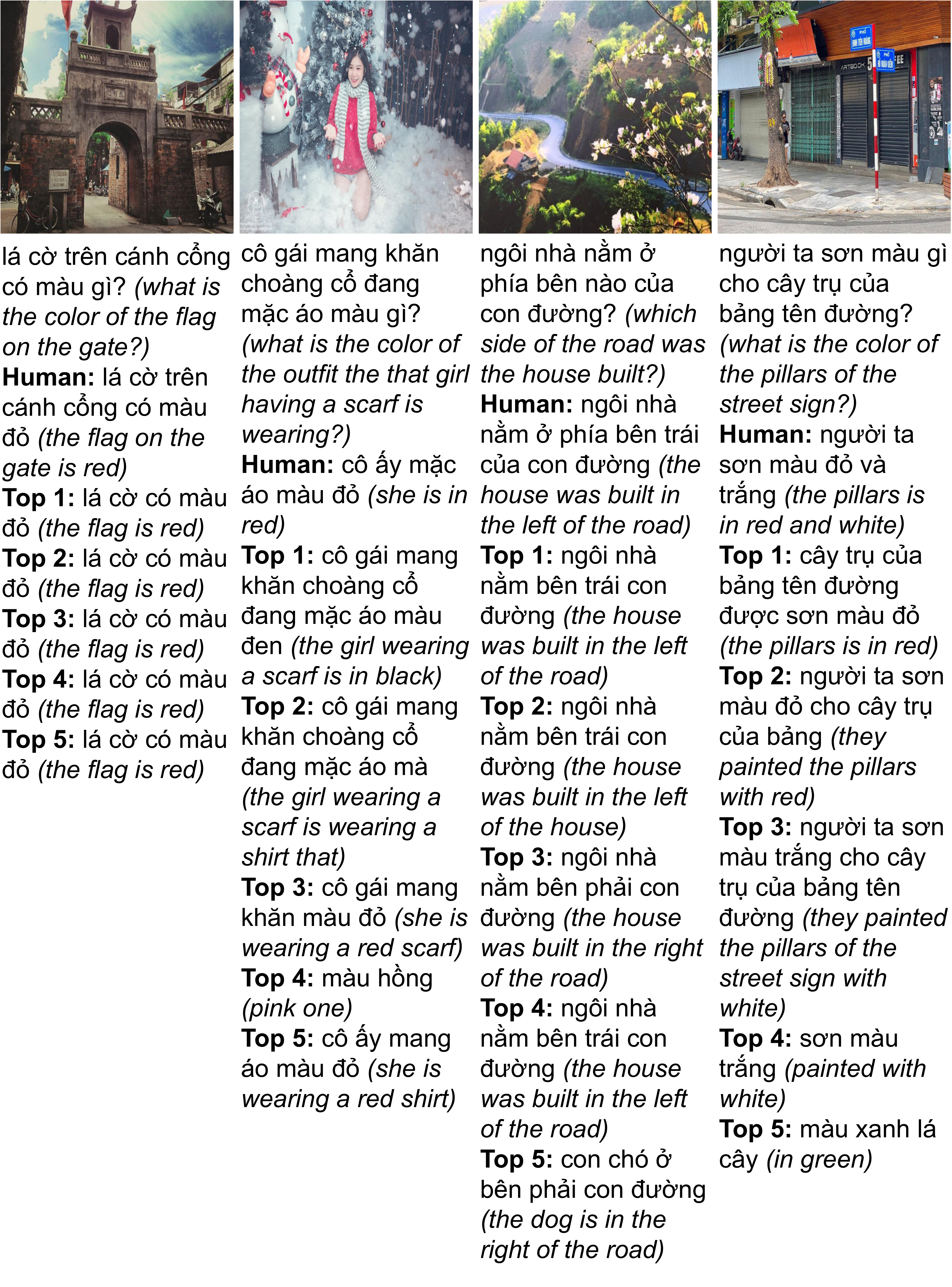}
    \caption{Results of top-5 models on medium answers in Vietnamese of the UIT-EVJVQA dataset}
    \label{fig:results_vi_medium-long}
\end{figure*}

\begin{figure*}
    \centering
    \includegraphics[width=0.88\textwidth]{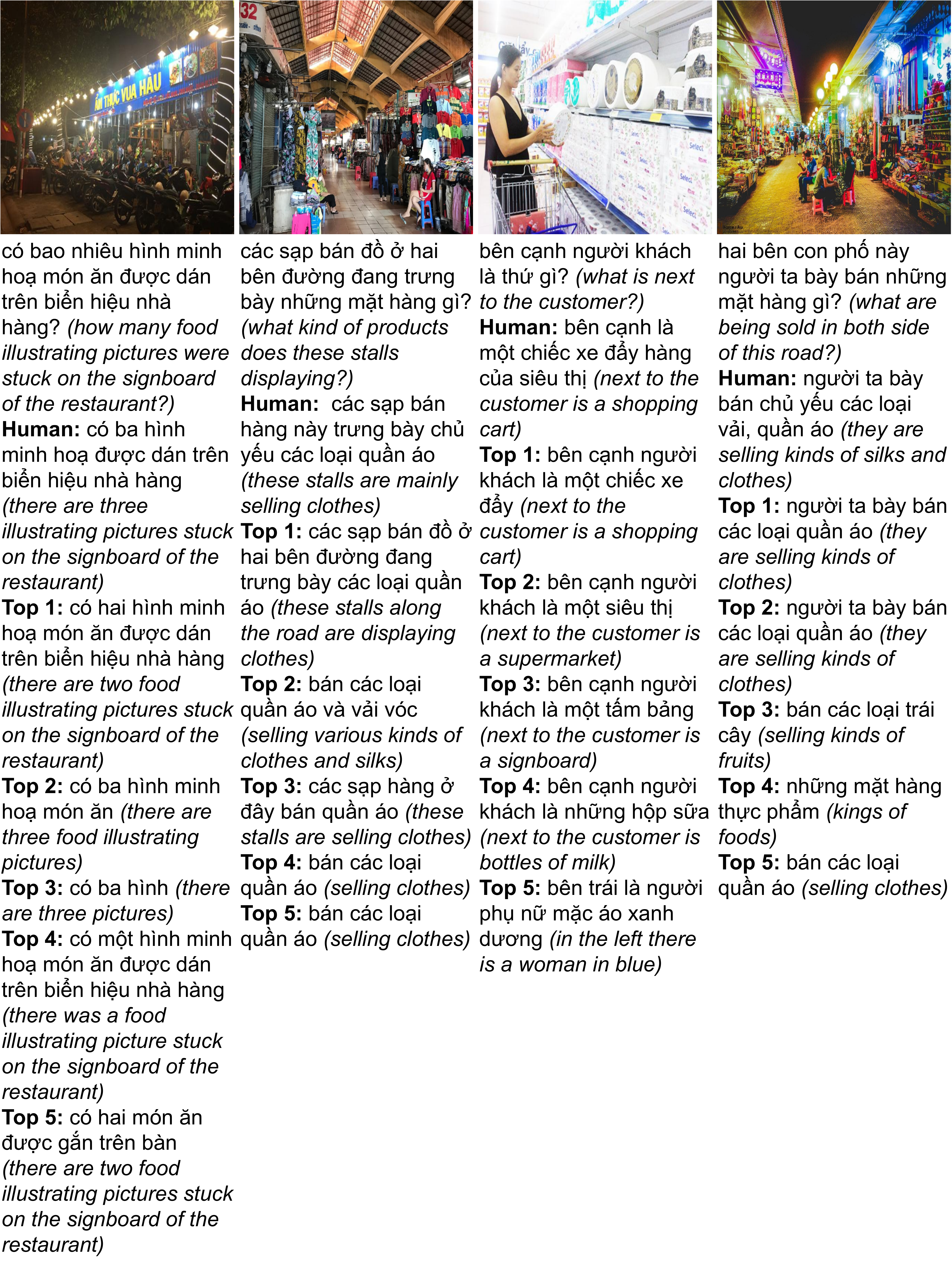}
    \caption{Results of top-5 models on long answers in Vietnamese of the UIT-EVJVQA dataset}
    \label{fig:my_label}
\end{figure*}

\begin{figure*}
    \centering
    \begin{subfigure}{0.9\textwidth}
        \centering
        \includegraphics[width=\textwidth]{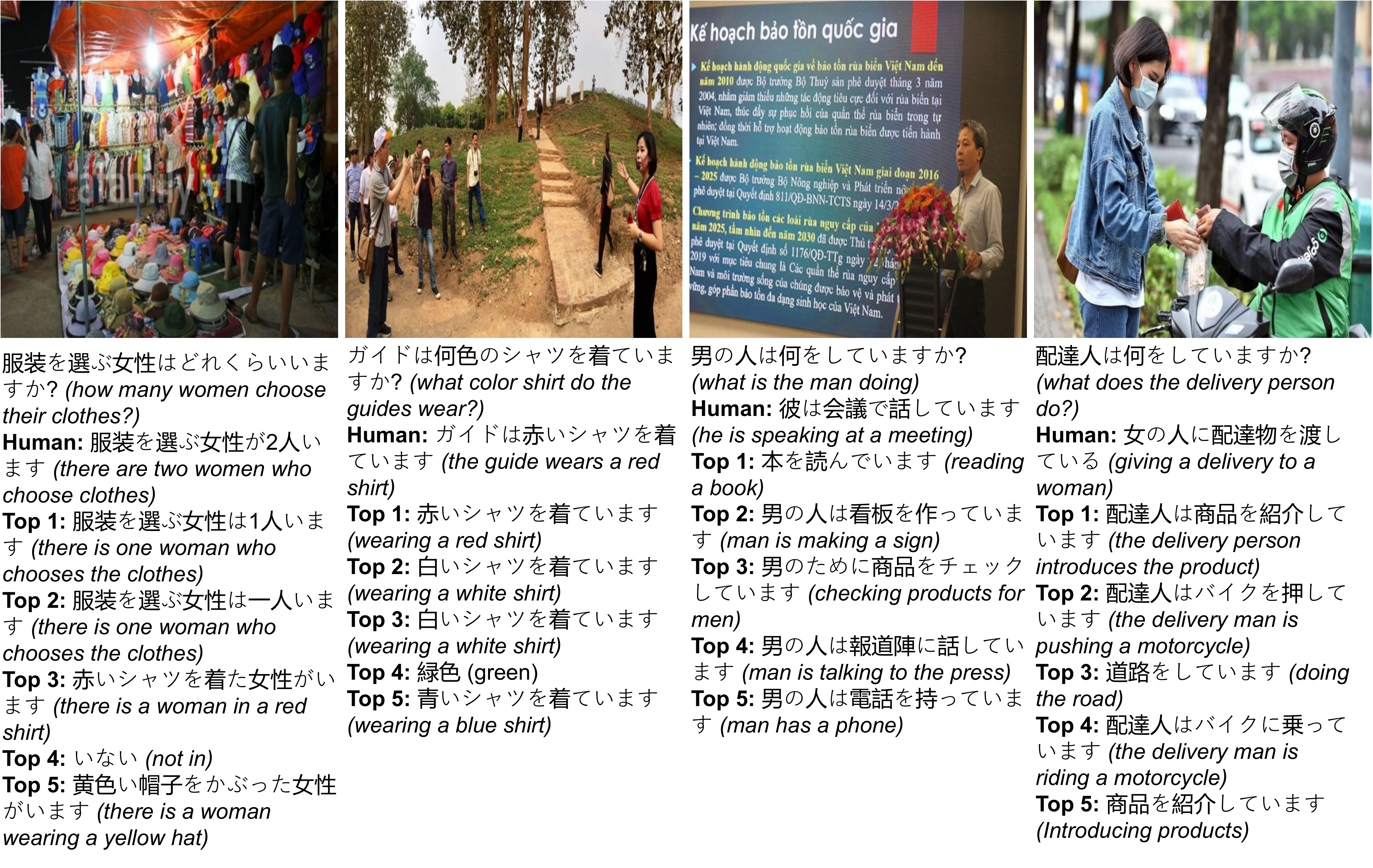}
    \end{subfigure}
    \begin{subfigure}{0.9\textwidth}
        \centering
        \includegraphics[width=\textwidth]{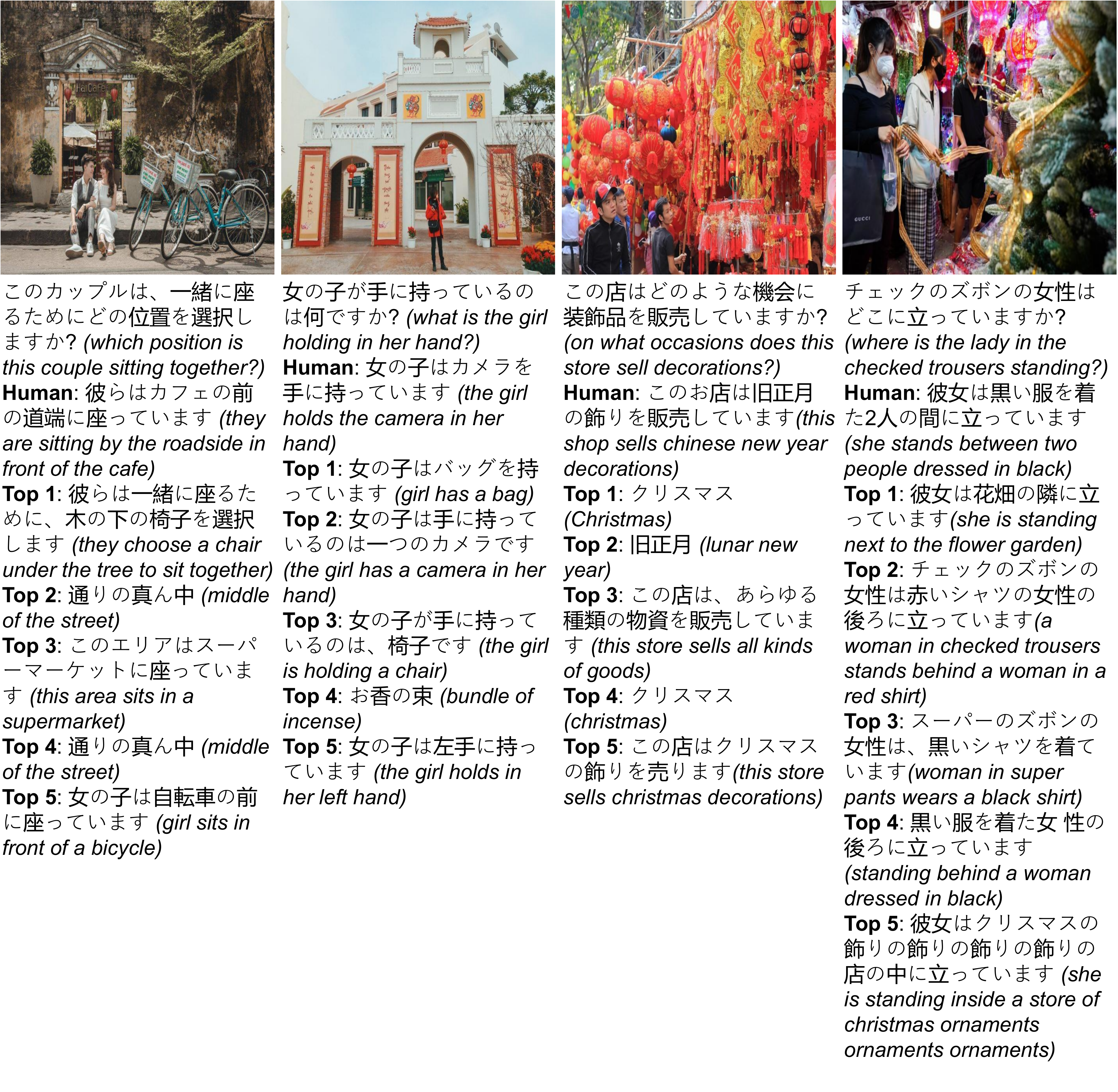}
    \end{subfigure}
    \caption{Results of top-5 models on long answers and very long answers in Japanese of the UIT-EVJVQA dataset}
    \label{fig:results_vi_long_very_long}
\end{figure*}